\documentclass[10pt,twocolumn,letterpaper]{article}

\usepackage{cvpr}              % To produce the CAMERA-READY version
\definecolor{cvprblue}{rgb}{0.21,0.49,0.74}
\usepackage[pagebackref,breaklinks,colorlinks,citecolor=cvprblue]{hyperref}

\usepackage[accsupp]{axessibility}
\usepackage{algorithm}%
\usepackage{algorithmicx}%
\usepackage{algpseudocode}%
\usepackage{colortbl}
\definecolor{secondcolor}{RGB}{189,215,238}
\definecolor{firstcolor}{RGB}{255,153,153}
\newcommand{\firstcolor}[1]{\cellcolor[rgb]{1,.60,.60}{#1}}
\newcommand{\secondcolor}[1]{\cellcolor[rgb]{.741,.843,.933}{#1}}

\newcommand{\bfsection}[1]{\vspace*{0.00cm}\noindent\textbf{#1.}}
\graphicspath{{Figure/}}
\def\paperID{4833} % *** Enter the Paper ID here
\def\confName{CVPR}
\def\confYear{2025}

\title{Task-driven Image Fusion with Learnable Fusion Loss}
\author{
        Haowen Bai$^{1}$\quad
        Jiangshe Zhang$^{1}$\thanks{Corresponding authors.}\quad
        Zixiang Zhao$^{2}$\footnotemark[1]\quad
        Yichen Wu$^{3}$\\
        Lilun Deng$^{1}$\quad
        Yukun Cui$^{1}$\quad
        Tao Feng$^{4}$\quad
        Shuang Xu$^{5}$\quad\\[1mm]
        $^{1}$Xi’an Jiaotong University\quad
        $^{2}$ETH Z\"urich\quad
        $^{3}$City University of Hong Kong\\
        $^{4}$Tsinghua University\quad
        $^{5}$Northwestern Polytechnical University\\
        {\tt\small hwbaii@stu.xjtu.edu.cn}
}
\begin{document}
\maketitle
\begin{abstract}
Multi-modal image fusion aggregates information from multiple sensor sources, achieving superior visual quality and perceptual features compared to single-source images, often improving downstream tasks.
However, current fusion methods for downstream tasks still use predefined fusion objectives that potentially mismatch the downstream tasks, limiting adaptive guidance and reducing model flexibility.
To address this, we propose Task-driven Image Fusion~\textbf{(TDFusion)}, a fusion framework incorporating a learnable fusion loss guided by task loss. 
Specifically, our fusion loss includes learnable parameters modeled by a neural network called the loss generation module. 
This module is supervised by the downstream task loss in a meta-learning manner.
The learning objective is to minimize the task loss of fused images after optimizing the fusion module with the fusion loss.
Iterative updates between the fusion module and the loss module ensure that the fusion network evolves toward minimizing task loss, guiding the fusion process toward the task objectives.
TDFusion’s training relies entirely on the downstream task loss, making it adaptable to any specific task.
It can be applied to any architecture of fusion and task networks.
Experiments demonstrate TDFusion’s performance through fusion experiments conducted on four different datasets, in addition to evaluations on semantic segmentation and object detection tasks.
The code is available at~\url{https://github.com/HaowenBai/TDFusion}.
\end{abstract}

\section{Introduction}
\label{sec:intro}
Multi-modal image fusion~\cite{tang2023datfuse,park2023cross,liu2021learning,zhao2023tufusion,li2023ccafusion, DBLP:journals/corr/abs-2211-14461, yang2021infrared} combines information from multiple sensors to produce a more holistic and detailed representation.
Infrared images capture thermal radiation regardless of lighting conditions, while visible images provide richer texture details.
Fused images enhance downstream tasks through improved information density and robustness~\cite{DBLP:journals/inffus/MaML19,DBLP:journals/tcsv/LiuCR20,DBLP:conf/aaai/JingLDWDSW20,DBLP:journals/tci/XuJWLSZZ20,DBLP:journals/tgrs/XuALZZL20, xiao2025event,xiao2024asymmetric}, outperforming single-modal inputs in semantic segmentation~\cite{DBLP:conf/iros/HaWKUH17,DBLP:journals/inffus/TangYM22,Liu_2023_ICCV,li2024object}, object detection~\cite{DBLP:journals/inffus/CaoGHYCQ19,DBLP:conf/cvpr/LiuFHWLZL22}, and other related applications~\cite{bai2024first,DBLP:conf/eccv/LiZHTW18,li2024ustc,albanwan2024image}. 
Conventional methods typically treat fusion as image restoration using unsupervised loss~\cite{zhao2023cddfuse,zhang2024mrfs,zhou2024probing,zhu2024task,DBLP:journals/tmm/ZhouWZML23} or perceptual loss~\cite{DBLP:journals/inffus/LiWK21,DBLP:journals/tip/LiW19,DBLP:journals/inffus/ZhangLSYZZ20}. 
These approaches prioritize visual-level fusion through predefined aggregation objectives, often neglecting semantic feature extraction.
This limitation hinders scene interpretation and task performance~\cite{DBLP:conf/iconip/HarisSU21,DBLP:conf/eccv/PeiHZLW18,DBLP:conf/cvpr/LiARWTJCZGC19,xiao2024event}. 
Recent advances explore the mutual enhancement between fusion and downstream tasks~\cite{liu2024infrared}.
By cascading the fusion network with downstream task network~\cite{ma2022toward,liu2022learning}, the task loss constrains the fusion learning, ensuring the fused images meet the task requirements~\cite{DBLP:journals/inffus/TangYM22,DBLP:conf/cvpr/LiuFHWLZL22}. Alternatively, some methods incorporate high-level visual task features~\cite{zhao2023metafusion,Liu_2023_ICCV,zhang2024mrfs} or focus on learning optimal initializations~\cite{liu2024task} to enhance fusion.

While integrating downstream tasks, existing frameworks still rely on \textit{predefined} fusion loss terms lacking dynamic adaptation.
The impact of downstream tasks remains limited due to specific combinations.
Manually defined losses preserve predefined guidance, frequently overlooking task-specific requirements.
This guidance imposes manually designed prior constraints on the fusion process, limiting the dynamic and adaptive influence of downstream tasks on specific image pairs.
These approaches, whether incorporating task features~\cite{zhao2023metafusion,Liu_2023_ICCV} or employing task losses~\cite{DBLP:journals/inffus/TangYM22,DBLP:conf/cvpr/LiuFHWLZL22}, still face the limitations of fixed fusion loss terms. 
Task-specialized networks~\cite{zhang2024mrfs} create fusion-task dependencies, restricting flexibility and limiting their applicability to various high-level vision tasks.
We address these limitations through a task-driven framework with learnable loss.
The fusion loss contains learnable parameters, generated by a loss generation module, and is designed to retain the intensity information of the source images for specific downstream tasks.
The purpose of updating the fusion loss is to guide the fusion network in generating fused images that minimize downstream task loss, thereby enhancing adaptability.
Moreover, the fusion loss update relies on the downstream task loss, making it independent of any specific task or network architecture.

The loss generation module produces fusion losses for subsequent fusion module updates.
This complexity presents challenges to standard end-to-end training.
Fortunately, meta-learning techniques, which are strategies for learning how to learn, can effectively achieve the learning objectives of the loss generation module. 
This involves minimizing task loss for fused images through an optimized fusion loss.
Meta-learning tackles common deep-learning challenges like limited data, high computational costs, and the need for better generalization. 
Core optimization areas include parameter tuning~\cite{finn2017model}, optimization strategies~\cite{li2017meta}, and network architectures search~\cite{liu2018darts}.
In this paper, we draw inspiration from the Model-Agnostic Meta-Learning (MAML) approach~\cite{finn2017model} to train our loss generation module. 
Specifically, training the loss generation module involves two stages: inner updates and outer updates. During inner updates, the output of the loss generation module updates a surrogate fusion module without altering the original parameters. 
During outer updates, the fused image from the surrogate module is fed into the task network. The resulting task loss then updates the loss generation module through backpropagation. 
This alternating training ensures that the loss generation module consistently produces fusion losses that minimize the downstream task loss of the fused images.

This paper introduces TDFusion, a task-oriented fusion framework driven by downstream tasks. It consists of a fusion module, a task module, and a loss generation module that learns to optimize the fusion loss. The fusion loss incorporates intensity preferences from source images and gradient preservation, guided by downstream task loss to refine intensity preferences. This model follows the general form of fusion loss used in advanced methods~\cite{zhao2023cddfuse, zhang2024mrfs, zhou2024probing, zhu2024task}, ensuring adaptability to various tasks. The updates of the loss generation module are performed using a meta-learning approach, optimizing the loss generation module parameters based on task loss from the fused images after each update of the fusion module. This process ensures that the loss function guides the fusion module continuously, optimizing feature aggregation and minimizing downstream task losses. The loss generation module dynamically adjusts through alternating updates with the fusion and task modules, generating optimal fusion losses at each model state.

Our contributions can be summarized as follows:
\begin{itemize}[itemsep=0cm,topsep=0cm,parsep=0pt]
\item We propose TDFusion, a meta-learning-based fusion framework that leverages the loss functions of downstream tasks for training.
This method promotes task-driven fusion and alleviates challenges caused by the absence of ground truth. 
Moreover, this framework is agnostic of specific downstream tasks or network architectures, which enhances its adaptability and flexibility.
\item Our framework includes a dynamically updated, learnable fusion loss generation module.
It selectively extracts source image information, minimizing the loss in downstream tasks.
This ensures optimal fusion performance while maximizing adaptability to downstream tasks.
\item We analyze the information preferences of downstream tasks such as semantic segmentation and object detection, providing deeper insights into multi-modal high-level vision tasks.
\item TDFusion achieves outstanding performance in both fusion and high-level vision tasks, validated on four fusion datasets with semantic segmentation and object detection.
\end{itemize}

\section{Related Work}
\subsection{Deep learning-based Image Fusion}
Deep learning-based image fusion methods have revolutionized the field by exploiting the powerful feature extraction capabilities of neural networks~\cite{liu2023bi,liu2023paif,liu2024searching,liu2022attention,liu2024coconet,liu2023holoco,bai2025retinex,DBLP:conf/aaai/Xu0LJG20,DBLP:journals/inffus/ZhangLSYZZ20,bai2024deep}.
These methods are broadly categorized into discriminative and generative approaches.
Discriminative methods~\cite{DBLP:journals/tcsv/ZhaoXZLZL22,DBLP:journals/pami/0002D21,DBLP:journals/corr/abs-2005-08448,zhou2024probing} leverage the strong reconstruction ability of neural networks to directly learn the mapping between source and fused images~\cite{DBLP:journals/tip/LiW19,DBLP:journals/inffus/LiWK21,DBLP:conf/ijcai/ZhaoXZLZL20,DBLP:journals/corr/abs-2211-14461}.
Generative methods, on the other hand, model the image generation process using generative approaches, integrating source images from a distributional perspective. 
These include methods based on Generative Adversarial Networks~\cite{DBLP:journals/inffus/MaYLLJ19,DBLP:journals/inffus/MaLYCGWJ20,DBLP:journals/tip/MaXJMZ20,DBLP:journals/tmm/ZhouWZML23} and diffusion models~\cite{DBLP:journals/corr/abs-2303-06840,yi2024diff}.
Unified fusion methods~\cite{9151265,DBLP:journals/ijcv/ZhangM21,zhu2024task} bridge the gap between different fusion sub-tasks, incorporating strategies such as continual learning~\cite{DBLP:journals/pami/XuMJGL22} and self-supervised decomposition techniques~\cite{DBLP:conf/eccv/LiangJLM22}.
The introduction of registration modules helps to mitigate misalignment issues~\cite{DBLP:conf/eccv/HuangLFLZL22,DBLP:conf/ijcai/WangLFL22,DBLP:conf/cvpr/Xu0YLL22} in source images.
Recent studies have further explored the synergy between fusion and high-level vision tasks. These include leveraging downstream task losses to optimize fusion networks~\cite{DBLP:journals/inffus/TangYM22,DBLP:conf/cvpr/LiuFHWLZL22}, embedding high-level task features~\cite{Liu_2023_ICCV,zhang2024mrfs,zhao2023metafusion}, and employing initialization techniques~\cite{liu2024task}.

\begin{figure*}[!]
	\centering
	\includegraphics[width=\linewidth]{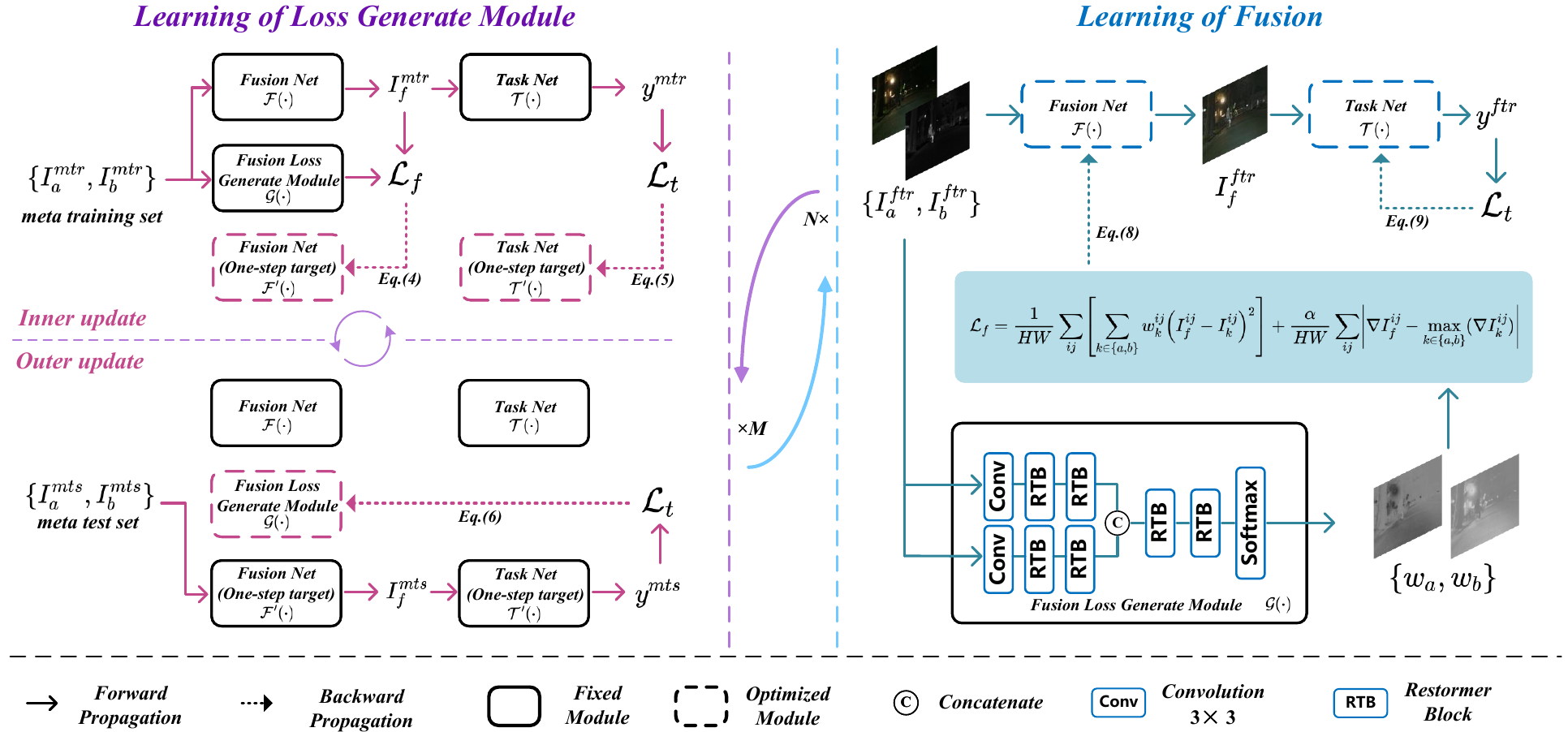}
        \vspace{-1.5em}
	\caption{{The TDFusion workflow alternates between training the loss generation module and the fusion module. Training of the loss generation module involves both inner and outer updates, learned through meta-learning.}}
        \vspace{-1.5em}
	\label{fig:workflow}
\end{figure*}

\subsection{Meta-Learning in Vision}
Meta-learning develops algorithms to automatically fine-tune hyperparameters for specific tasks, showing its versatility and effectiveness across various domains. MAML~\cite{finn2017model} and its variations~\cite{finn2019online,DBLP:journals/corr/abs-1803-02999,DBLP:conf/cvpr/QinQPJ23} focus on learning efficient initialization parameters to quickly adapt to new tasks using minimal data. Meta-SGD~\cite{li2017meta} extends MAML by learning optimal update directions and rates, beneficial in few-shot learning scenarios. Other approaches like MW-Net~\cite{shu2019meta} and L2RW~\cite{ren2018learning} emphasize selecting relevant sample weights to tackle noisy data using a compact validation set. Additionally, some studies focus on improving model adaptability through learning loss functions~\cite{DBLP:conf/nips/HouthooftCISWHA18,DBLP:conf/nips/AntoniouS19,DBLP:conf/iccv/BaikCKCML21}.

In image fusion, learnable filters~\cite{DBLP:journals/tip/LiCLCY21} enable the fusion of images at arbitrary resolutions. MetaFusion~\citep{zhao2023metafusion} introduces a mechanism that improves image fusion and object detection by aligning semantics with fusion-specific features. ReFusion~\cite{bai2024refusion} guides the learnable fusion loss for various fusion tasks by reconstructing the source images.
Meta-learning also supports neural architecture search~\cite{DBLP:conf/mm/LiuLL021,liu2024task} to identify optimal network architectures for image fusion and customizes network initialization for various tasks~\cite{liu2024task}.

\subsection{Comparison with Existing Approaches}
We propose a novel image fusion method tailored for downstream tasks, leveraging a learnable fusion loss driven by task-specific objectives. Our method employs a meta-learning algorithm that alternates between inner and outer updates, allowing the downstream task loss to guide the optimization of learnable fusion parameters. This results in fused images that minimize the downstream task loss, enhancing their adaptability across various tasks. Unlike previous methods, our approach develops a task-specific fusion loss, shifting focus from traditional factors such as resolution and network structure, and avoiding reliance on predefined fusion loss terms. This renders our fusion framework more flexible and applicable to various scenarios.

\section{Method}
\subsection{Overview}
Our TDFusion framework, as shown in~\cref{fig:workflow}, consists of a fusion network $\mathcal{F}(\cdot)$, a downstream task network $\mathcal{T}(\cdot)$, and a fusion loss generation module $\mathcal{G}(\cdot)$, which produces parameters for a learnable loss function.
The parameters of these modules are denoted as $\theta_\mathcal{F}$, $\theta_\mathcal{T}$, and $\theta_\mathcal{G}$, respectively. 
The one-step updated clones of $\mathcal{F}$ and $\mathcal{T}$ are denoted as $\mathcal{F'}$ and $\mathcal{T'}$, with parameters $\theta_\mathcal{F'}$ and $\theta_\mathcal{T'}$.
During the updates, the fusion network and the loss generation module alternate in learning, as depicted by blue and purple in \cref{fig:workflow}.
The update of the loss generation module consists of inner and outer updates, detailed in the following subsections. $\mathcal{L}_f$ and $\mathcal{L}_t$ represent the learnable fusion loss and task-specific loss, with their formulations provided in the next section.

\subsection{Loss Function}

The learnable fusion loss $\mathcal{L}_f$ consists of the intensity term and the gradient term. 
The intensity term is defined by the output of the loss generation module $\{w_{a}, w_{b}\} =\mathcal{G}(I_{a}, I_{b})$, where $w_{a}$ and $w_{b}$ control the intensity preference in the fusion loss. 
The Softmax function in the loss generation module ensures $w_{a}^{ij} + w_{b}^{ij} = 1$ for each pixel, thereby selectively retaining the intensity information from the source images. 
The gradient term emphasizes higher gradient values from the input images~\cite{zhao2023cddfuse,zhang2024mrfs,zhu2024task}, aiming to preserve maximal information from the source. 
The detailed formulation of the learnable fusion loss is as follows:
\begin{align}
\small
	\mathcal{L}_f&=\mathcal{L}_f^{int}+\alpha \mathcal{L}_f^{grad},\\
	\mathcal{L}_f^{int}&=\frac{1}{H W} \sum_{i j}\left[\sum_{k \in\{a, b\}} w_k^{ij}\left(I_f^{i j}-I_k^{i j}\right)^2\right],\label{eq2}\\
	\mathcal{L}_f^{grad}&=\frac{1}{H W} \sum_{i j}\left|\nabla I_f^{i j}- \max_{k \in\{a, b\}}(\nabla I_k^{i j}) \right|,
\end{align}
where $\nabla$ denotes the Sobel operator, commonly employed for gradient extraction in image fusion~\cite{zhao2023cddfuse,DBLP:conf/cvpr/LiuFHWLZL22,tang2022image}. The parameter $\alpha$ serves as a scaling factor, while $\mathcal{L}_f^{int}$ and $\mathcal{L}_f^{grad}$ represent the intensity loss and gradient loss, respectively.
The weights $\{w_{a}, w_{b}\}$ control the emphasis of the loss function on the intensity information from each source image.
These parameters enable the fusion process to selectively aggregate and incorporate relevant information from the source images.
Variations in $\theta_\mathcal{G}$ lead to different configurations of $w_a$ and $w_b$, thereby influencing the characteristics of the fusion loss.
$\theta_\mathcal{G}$ undergoes updates driven by the high-level task loss during the in-step update process, as detailed in \cref{sec:LGM}.

The loss function $\mathcal{L}_t$ depends on the specific task.
In this study, we adopt SegFormer~\cite{cheng2021segformer} and YOLOv8~\cite{ultralytics_yolov8_2023} for the downstream tasks, employing cross-entropy loss and YOLO loss~\cite{ultralytics_yolov8_2023} for each task, respectively.

\subsection{Dataset Partitioning}
In order to enhance the effectiveness of the loss generation module $\mathcal{G}$ in guiding the fusion tasks, we create non-overlapping subsets of size $\emph{M}$ at each training epoch. 
The meta-training set $\{I_a^{mtr}, I_b^{mtr}\}$ and the meta-test set $\{I_a^{mts}, I_b^{mts}\}$ are randomly drawn from the fusion training set $\{I_a^{ftr}, I_b^{ftr}\}$. 
These subsets are fed into the model sequentially during the training process of the loss generation module, covering both the inner and outer updates.
The entire fusion training set $\{I_a^{ftr}, I_b^{ftr}\}$ is fully utilized during the training of the fusion network.

\subsection{Learning of loss generation module}\label{sec:LGM}
\cref{fig:workflow} illustrates that the loss generation and fusion modules are trained alternately, ensuring the fusion loss is optimized at various stages of training and under different states of the fusion network.
The process of training to optimize the fusion loss involves two key steps: the inner update and the outer update.
In the inner update, clones of both the fusion network and the task network are generated, with each network undergoing a single training iteration using the fusion loss and task loss, respectively.
This process is designed to obtain the state of the network guided by the fusion loss.
During the outer update, the task loss of the fusion image produced by the updated clone is calculated. This loss is then backpropagated to the loss generation network.
The goal of this step is to direct the fusion network to generate fusion images that result in lower downstream task loss, once guided by the fusion loss.
The alternating updates between the inner and outer steps constitute the learning procedure for the loss generation module.

\subsubsection{Inner Update}
During the inner update phase, the fusion network $\mathcal{F}$ undergoes a single update guided by the fusion loss, which depends on the current state of network $\mathcal{G}$.
This update primarily aims to compute the intermediate parameters $\theta_\mathcal{F'}$ and $\theta_\mathcal{R'}$, which are crucial for updating $\theta_\mathcal{G}$ in the subsequent phase. 
The upper section of the purple region in~\cref{fig:workflow} illustrates this process.
During this phase, the images from the meta-training set $\{I_a^{mtr}, I_b^{mtr}\}$ are fed into the model:
\begin{equation}
	\label{equ1}
	\small
	\theta_{\mathcal{F'}}=\theta_\mathcal{F}-\eta_\mathcal{F'} \frac{\partial \mathcal{L}_f\left(I_a^{mtr}, I_b^{mtr}, I_f^{mtr} ; \theta_\mathcal{G}\right)}{\partial \theta_\mathcal{F}},
\end{equation}
where $\mathcal{F}$ undergoes a single gradient descent update.
$\theta_\mathcal{G}$ represent the parameters of the loss generation module $\mathcal{G}$, which determine the parameters for the learnable fusion loss.
And the notation $\eta_\mathcal{F'}$ refers to the step size.
The module $\mathcal{F'}$ temporarily substitutes $\mathcal{F}$, adjusting its parameters in one update step.
Meanwhile, the parameters of $\mathcal{F}$, denoted as $\theta_\mathcal{F}$, remain unchanged.
Similarly, $\mathcal{T'}$ is updated in a single step using the parameters $\theta_\mathcal{T}$ from the current task network $\mathcal{T}$:
\begin{equation}\label{equ2}
	\small
	\theta_{\mathcal{T'}}=\theta_\mathcal{T}-\eta_\mathcal{T'} \frac{\partial \mathcal{L}_t\left(I_f^{mtr}\right)}{\partial \theta_\mathcal{T}}.
\end{equation}
The parameters $\theta_{\mathcal{F'}}$ and $\theta_{\mathcal{T'}}$ are both updated during the inner update, which also ensures that the computation graph of $\theta_{\mathcal{F'}}$ with respect to $\theta_{\mathcal{G}}$ is preserved. This preserved graph is essential for optimizing $\theta_{\mathcal{G}}$ during the outer update.
\subsubsection{Outer Update}
The primary objective of the outer update is to evaluate and refine the fusion guidance capability of $\mathcal{G}$, specifically by strengthening the influence of the loss function $\mathcal{L}_f$ in steering the fusion module $\mathcal{F}$.
In the framework diagram, this stage is represented in the lower part of the purple region.
The modules $\mathcal{F'}$ and $\mathcal{T'}$, derived from the inner update, represent the current guidance capacity of $\mathcal{G}$.
In an ideal scenario, the optimal fusion loss should enhance the performance of the downstream task on the fused image.
In this stage, the meta-test set $\{I_a^{mts}, I_b^{mts}\}$ is employed.
The parameters $\theta_{\mathcal{G}}$ are subsequently updated using the task loss $\mathcal{L}_t$, {which is computed by $\mathcal{F'}$ and $\mathcal{T'}$:}
\begin{equation}\label{equ3}
	\small
\theta_\mathcal{G}=\theta_\mathcal{G}-\eta_\mathcal{G} \frac{\partial \mathcal{L}_t\left(I_f^{mts}\right)}{\partial \theta_\mathcal{G}},
\end{equation}
where $I_f^{mts}=\mathcal{F'}(I_a^{mts},I_b^{mts})$, and the gradient $\partial \mathcal{L}_t/\partial \theta_\mathcal{G}$ can be calculated as:
\begin{equation}
	\small
	\frac{\partial \mathcal{L}_t}{\partial \theta_\mathcal{G}}=\frac{\partial \mathcal{L}_t}{\partial \theta_\mathcal{F'}} \!*\!\left(-\eta_\mathcal{F'} \frac{\partial^2 \mathcal{L}_f\left(I_a^{mtr}, I_b^{mtr}, I_f^{mtr} ; \theta_\mathcal{G}\right)}{\partial \theta_\mathcal{F} \partial \theta_\mathcal{G}}\right).
\end{equation}
\cref{equ3} holds because the task loss $\mathcal{L}_t$ is determined by $I_f^{mts}$, which in turn relies on $\theta_\mathcal{F'}$.
The optimization of $\theta_\mathcal{G}$ via $\mathcal{L}_t$ is realized by preserving the computational relationship between $\theta_\mathcal{F'}$ and $\theta_\mathcal{G}$ throughout the inner update.
The updated $\mathcal{G}$ module gains the ability to generate enhanced fusion loss functions, enabling the fusion module to integrate relevant information from the source images more efficiently into the fused output for downstream tasks.

\subsection{Learning of Fusion Network}
The alternating inner and outer update iterations form a flexible and effective mechanism to refine $\mathcal{G}$ in response to the evolving state of $\mathcal{F}$.
After refining $\mathcal{G}$, it is utilized to further improve the training of $\mathcal{F}$.
This stage, denoted in blue in the diagrams, involves processing the fusion training set images $\{I_a^{ftr}, I_b^{ftr}\}$.
Both $\mathcal{F}$ and $\mathcal{T}$ are updated through the application of the fusion loss $\mathcal{L}_f$ and {task loss $\mathcal{L}_t$:}
\begin{equation}
\small
	\theta_{\mathcal{F}}=\theta_\mathcal{F}-\eta_\mathcal{F} \frac{\partial \mathcal{L}_f\left(I_a^{ftr}, I_b^{ftr}, I_f^{ftr} ;\theta_{\mathcal{G}}\right)}{\partial \theta_\mathcal{F}},\label{equ4}
\end{equation}
\begin{equation}
\small
	\theta_{\mathcal{T}}=\theta_\mathcal{T}-\eta_\mathcal{T} \frac{\partial \mathcal{L}_t\left(I_f^{ftr}\right)}{\partial \theta_\mathcal{T}}.\label{equ5}
\end{equation}
After completing several training sessions on the fusion network, the focus then shifts back to the learning phase of the fusion generation module.
The fusion framework evolves through a series of alternating phases, with each phase fine-tuning the fusion loss based on the fusion network's current state.
Such alternating phases ensure that the fusion network consistently applies the most suitable fusion loss during its progression. 
Ultimately, this results in the development of a highly efficient fusion network, optimized for peak performance. 
The complete training procedure is detailed in~\cref{alg}.

\begin{algorithm}[t]
	\caption{TDFusion Training Algorithm}
	\label{alg}
	\begin{algorithmic}[1]
	\Statex\textbf{Require:}
		{Training set $\{I_a^{ftr}, I_b^{ftr}\}$ with size $\emph{N}$.}
	\Statex\textbf{Output:}  
		{Thoroughly trained $\theta_{\mathcal{F}}$, $\theta_{\mathcal{T}}$, $\theta_{\mathcal{G}}$.}
		\State Initialize $\theta_{\mathcal{F}}$, $\theta_{\mathcal{T}}$, $\theta_{\mathcal{G}}$.
		\For {$epoch=1$ \text{\textbf{to}} $\textbf{\emph{L}}$}
		\State Sample $\{I_a^{mtr}, I_b^{mtr}\}$ and $\{I_a^{mts}, I_b^{mts}\}$.
		\For {$step=1$ \text{\textbf{to}} $\textbf{\emph{M}}$}
		\Statex \qquad \quad \% \textit{Inner update: apply $\mathcal{G}$}.
		\State Sample $(I_a^{mtr}, I_b^{mtr})$ and get $(I_f^{mtr}, y^{mtr})$. 
		\State Compute $\theta_{\mathcal{F'}}$ and $\theta_{\mathcal{T'}}$ by~\cref{equ1} and ~\cref{equ2}.
		\Statex \qquad \quad \% \textit{Outer update: optimize $\mathcal{G}$}.
		\State Sample $(I_a^{mts}, I_b^{mts})$ and get $(I_f^{mts}, y^{mts})$. 
		\State Update $\theta_{\mathcal{G}}$ by~\cref{equ3}.
		\EndFor
		\For {$step=1$ \text{\textbf{to}} $\textbf{\emph{N}}$}
		\Statex \qquad \quad \% \textit{Fusion update: optimize $\mathcal{F}$ and $\mathcal{T}$}.
		\State Sample $(I_a^{ftr}, I_b^{ftr})$ and get $(I_f^{ftr},y^{ftr})$. 
		\State Update $\theta_{\mathcal{F}}$ and $\theta_{\mathcal{T}}$ by~\cref{equ4} and~\cref{equ5}.
		\EndFor
		\EndFor      
        
	\end{algorithmic}
\end{algorithm}

\subsection{Network Architecture}
TDFusion is composed of three modules: the fusion network, the downstream task network, and the loss generation network. 
The fusion network shares the same architecture as~\cite{bai2024refusion}, a lightweight model built upon the Restormer Block (RTB)~\citep{zamir2022restormer}.
An adaptive fusion module is incorporated into this network to facilitate feature integration.
\cref{fig:workflow} illustrates the structure of the loss generation module.
It also employs the Restormer Block (RTB)~\citep{zamir2022restormer} as its primary component, receiving inputs $\{I_a,I_b\}$.
After applying $Softmax(\cdot)$, the final output guarantees $w_{a}^{ij}+w_{b}^{ij}=1$ for each pixel.
This design ensures that the fused image meets the similarity constraints and removes the reliance on initialization within the loss generation module.
The architecture of the downstream task network $\mathcal{T}(\cdot)$ depends on the specific task.
For learning of loss generation module, we chose the most lightweight models of SegFormer~\cite{cheng2021segformer} and YOLOv8~\cite{ultralytics_yolov8_2023} for semantic segmentation and object detection, respectively.

\subsection{Theoretical Analysis}

To better understand the weighting mechanism of the loss generation module $\mathcal{G}$, we investigate the optimization procedure of $\mathcal{G}$, which generates the weights $\{w_a, w_b\}$, denoted as $\theta_{\mathcal{G}}$. For clarity, we rewrite~\cref{eq2} as follows:
\begin{equation}
\small
	\begin{aligned}
		\mathcal{L}_f^{int} = &[{w}_a\odot({I}_a-{I}_f)\odot({I}_a-{I}_f)   \\
		&+ {w}_b\odot({I}_b-{I}_f)\odot({I}_b-{I}_f) ] \times \frac{1}{HW} \\
		= & [\mathcal{G}({I}_a,{I}_b;\theta_{\mathcal{G}})  \odot({I}_a-\mathcal{F}_{\theta_{\mathcal{F}}}({I}_a,{I}_b)) 
		\\ &\quad\odot({I}_a-\mathcal{F}_{\theta_{\mathcal{F}}}({I}_a,{I}_b))  \\
		&  +(1-\mathcal{G}({I}_a,{I}_b;\theta_{\mathcal{G}}))\odot({I}_b-\mathcal{F}_{\theta_{\mathcal{F}}}({I}_a,{I}_b)) \\ & \quad\odot({I}_b-\mathcal{F}_{\theta_{\mathcal{F}}}({I}_a,{I}_b)) ] \times \frac{1}{HW}. 
	\end{aligned}
\end{equation}

\begin{figure*}[t]
	\centering
	\includegraphics[width=\linewidth]{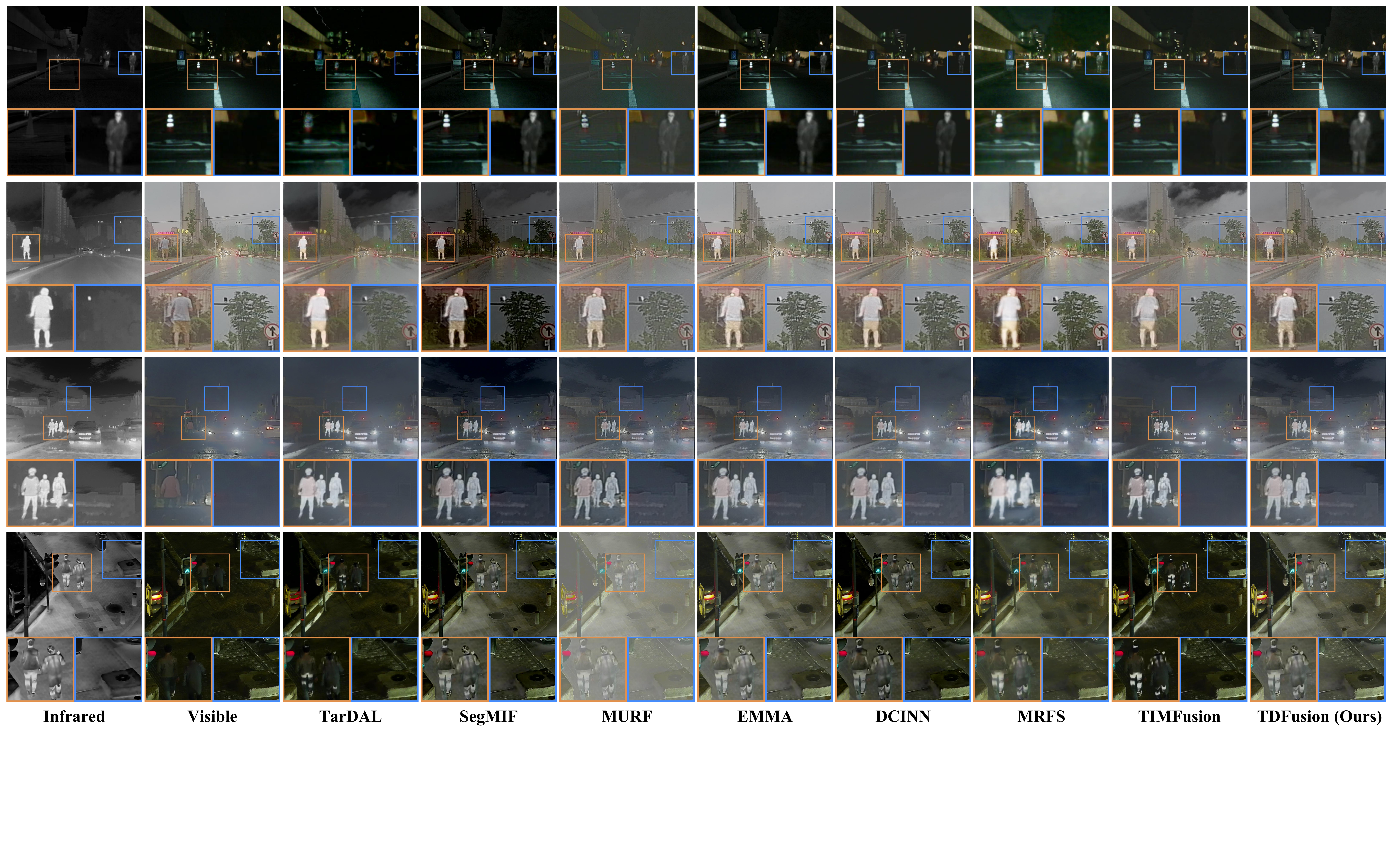}
        \vspace{-2em}
	\caption{{Visual comparison of fusion results. The cases are “01258N” in MSRS dataset, “00122” in FMB dataset, “00449” in M3FD dataset and “200304” in LLVIP dataset.}}
    \vspace{-1.5em} 
	\label{fig:fusion}
\end{figure*}

Here, $w_a, w_b \in \mathbb{R}^{H \times W}$, 
$I_a, I_b \in \mathbb{R}^{H \times W}$, and $\odot$ denotes the element-wise multiplication operation. Let $\Omega^{'}$ be the set $\{\theta_{\mathcal{F'}}, \theta_{\mathcal{T'}}\}$, this leads to the following expression:
\begin{equation}
    \small
	\begin{aligned}
		\theta_{\mathcal{G}} =& \theta_{\mathcal{G}} -\eta_{\mathcal{G}} \frac{\partial \mathcal{L}_t^{mts}(\Omega^{'}(\theta_{\mathcal{G}}))}{\partial \theta_{\mathcal{G}}} \\
		= &\theta_{\mathcal{G}} - \eta_{\mathcal{G}} \frac{\partial \mathcal{L}_t^{mts}(\Omega^{'}(\theta_{\mathcal{G}}))} {\partial \Omega^{'}}  \frac{\partial \Omega^{'}(\theta_{\mathcal{G}})}{\partial \theta_{\mathcal{G}}} \\
		= & \theta_{\mathcal{G}} - \eta_{\mathcal{G}}\eta_{\mathcal{F'}} \underbrace{\frac{\partial \mathcal{L}_t^{mts}(\Omega^{'}(\theta_{\mathcal{G}}))} {\partial \Omega^{'}}}_{(\mathrm{a})} \times  \frac{\partial \mathcal{G}({I}_a,{I}_b;\theta_{\mathcal{G}})}{\theta_{\mathcal{G}}} \\& \times\underbrace{[({I}_a - \frac{\partial \mathcal{F}_{\theta_{\mathcal{F}}}}{\partial{\theta_{\mathcal{F}}}})\odot({I}_a - \frac{\partial \mathcal{F}_{\theta_{\mathcal{F}}}}{\partial{\theta_{\mathcal{F}}}})}_{(\mathrm{b})} \\ & \quad \underbrace{-({I}_b - \frac{\partial \mathcal{F}_{\theta_{\mathcal{F}}}}{\partial{\theta_{\mathcal{F}}}})\odot({I}_b - \frac{\partial \mathcal{F}_{\theta_{\mathcal{F}}}}{\partial{\theta_{\mathcal{F}}}}) ]}_{(\mathrm{b})} \\
		= & \theta_{\mathcal{G}}-\eta_{\mathcal{G}}\eta_{\mathcal{F}^{'}} \mathbf{G}\times \frac{\partial \mathcal{G}({I}_a,{I}_b;\theta_{\mathcal{G}})}{\theta_{\mathcal{G}}}.
	\end{aligned}
\end{equation}
Here, $\mathbf{G}$ denotes the inner product between two gradients: (a) the first one is derived from the task loss using a \textbf{meta-testing set}, and (b) the second is calculated from the fusion loss based on a \textbf{meta-training set}.
Consequently, the optimization of the module $\theta_{\mathcal{G}}$ is driven by the task loss, with the objective of preserving task-specific information throughout the fusion process.

\section{Experiment}
\subsection{Setup}
\bfsection{Experimental Setup}
In our experiments, the epoch number $\textbf{\emph{L}}$ and the training iterations of the loss generation module $\textbf{\emph{M}}$ are set to 50 and 200. The learning iterations for the fusion network $\textbf{\emph{N}}$ depend on the size of the dataset. 
We use the Adam optimizer with a learning rate of 1e-4, a batch size of 2, and a hyperparameter $\alpha$ set to 1. 
All experiments are conducted on a PC with a single NVIDIA RTX 3090 GPU.

\bfsection{Evaluation Metrics and Comparison Methods}
The advanced fusion methods compared in our study include TarDAL~\cite{DBLP:conf/cvpr/LiuFHWLZL22}, SegMIF~\cite{Liu_2023_ICCV}, MURF~\cite{xu2023murf}, EMMA~\cite{zhao2024equivariant}, DCINN~\cite{wang2024general}, MRFS~\cite{zhang2024mrfs}, and TIMFusion~\cite{liu2024task}. 
The fusion performance is evaluated using metrics including entropy (EN), spatial frequency (SF), sum of correlation differences (SCD), visual information fidelity (VIF), $Q^{AB/F}$, and structural similarity index metric (SSIM).

\bfsection{Dataset Split}
We use four datasets annotated for downstream tasks, including MSRS~\cite{DBLP:journals/inffus/TangYZJM22}, FMB~\cite{Liu_2023_ICCV}, M3FD~\cite{DBLP:conf/cvpr/LiuFHWLZL22}, and LLVIP~\cite{jia2021llvip}.
MSRS contains 1083/361 image pairs for training/test, and FMB contains 1220/280 pairs for training/test. We follow the splits of the original papers for both datasets.
M3FD dataset consists of 4200 pairs of images for detection, with 300 pairs designated for fusion evaluation. The 4200 detection images are split into 3150 for training and 1050 for testing, ensuring that the 300 pairs for fusion evaluation are included within the detection test set. These 300 fusion images are then employed to assess the fusion performance.
The original LLVIP dataset contains 12025/3463 image pairs as training/test set. Due to its large size, we select every 10th image to form our training and test sets, resulting in 1203/347 image pairs for training and testing. Our splits for M3FD and LLVIP will be available.

\subsection{Fusion Experiments}

\begin{table*}[t]
	\centering
	\caption{Quantitative comparison of Infrared-visible image fusion. The \colorbox{firstcolor}{red} and \colorbox{secondcolor}{blue} markers represent the best and second-best values.}
    \vspace{-0.5em}
	\label{tab:fusion}%
	\resizebox{\linewidth}{!}{
		\begin{tabular}{lcccccclcccccc}
			\toprule
			\multicolumn{7}{c}{\textbf{Infrared-visible Image Fusion on MSRS~\cite{DBLP:journals/inffus/TangYZJM22} Dataset}}                              &                              \multicolumn{7}{c}{\textbf{Infrared-visible Image Fusion on FMB~\cite{Liu_2023_ICCV} Dataset}}                               \\
			&  EN $\uparrow$   &   SF $\uparrow$   &  SCD $\uparrow$   &  VIF $\uparrow$  & $Q^{AB/F}$$\uparrow$  & SSIM  $\uparrow$  &                                        &  EN $\uparrow$   &   SF $\uparrow$   &  SCD $\uparrow$   &  VIF $\uparrow$  & $Q^{AB/F}$$\uparrow$  & SSIM  $\uparrow$   \\ \midrule
            TarDAL~\cite{DBLP:conf/cvpr/LiuFHWLZL22}    & 5.28 & 5.98  & 0.71 & 0.21 & 0.18 & 0.47 & TarDAL~\cite{DBLP:conf/cvpr/LiuFHWLZL22} & 6.63 & 6.94  & 1.03 & 0.28 & 0.29 &\secondcolor{0.74} \\
            SegMIF~\cite{Liu_2023_ICCV}    & 5.95 & 11.10 & 1.57 & 0.44 & 0.63 & 0.55 & SegMIF~\cite{Liu_2023_ICCV} &\secondcolor{6.83} & 13.69 &\secondcolor{1.72} & 0.39 & 0.65 & 0.60\\
            MURF~\cite{xu2023murf}      & 5.04 & 10.49 & 1.02 & 0.22 & 0.37 & 0.60 & MURF~\cite{xu2023murf} & 6.37 & 13.88 & 1.34 & 0.22 & 0.37 & 0.68\\
            EMMA~\cite{zhao2024equivariant}      & 6.73 & \firstcolor{11.56} &\secondcolor{1.62} &\secondcolor{0.49} &\secondcolor{0.64} &\secondcolor{0.70} & EMMA~\cite{zhao2024equivariant} & 6.77 & \firstcolor{15.00} & 1.50 &\secondcolor{0.42} &\secondcolor{0.65} & 0.72\\

            DCINN~\cite{wang2024general}     & 6.00 & 10.51 & 1.49 & 0.41 & 0.57 & 0.52 & DCINN~\cite{wang2024general} & 6.47 & 11.47 & 1.39 & 0.38 & 0.59 & 0.74\\
            MRFS~\cite{zhang2024mrfs}      & \firstcolor{7.00} & 8.86  & 1.42 & 0.37 & 0.49 & 0.55 & MRFS~\cite{zhang2024mrfs} & 6.78 & 12.42 & 1.24 & 0.38 & 0.62 & 0.73\\
            TIMFusion~\cite{liu2024task} & 6.27 & 9.67  & 1.34 & 0.32 & 0.48 & {0.68} & TIMFusion~\cite{liu2024task} & 6.51 & 12.23 & 1.24 & 0.35 & 0.59 & 0.73\\
            TDFusion (Ours)  &\secondcolor{6.74} &\secondcolor{11.30} & \firstcolor{1.86} & \firstcolor{0.50}& \firstcolor{0.67} & \firstcolor{0.70} & TDFusion (Ours)  & \firstcolor{6.86} &\secondcolor{14.16} & \firstcolor{1.76} & \firstcolor{0.43} & \firstcolor{0.68} & \firstcolor{0.75}\\
			\midrule
			\multicolumn{7}{c}{\textbf{Infrared-visible Image Fusion on M3FD~\cite{DBLP:conf/cvpr/LiuFHWLZL22} Dataset}}                                      &                                            \multicolumn{7}{c}{\textbf{Infrared-visible Image Fusion on LLVIP~\cite{jia2021llvip} Dataset}}                                            \\
			&  EN $\uparrow$   &   SF $\uparrow$   &  SCD $\uparrow$   &  VIF $\uparrow$  & $Q^{AB/F}$$\uparrow$  & SSIM  $\uparrow$  &                                        &  EN $\uparrow$   &   SF $\uparrow$   &  SCD $\uparrow$   &  VIF $\uparrow$  & $Q^{AB/F}$$\uparrow$  & SSIM  $\uparrow$
            \\ \midrule
            TarDAL~\cite{DBLP:conf/cvpr/LiuFHWLZL22}    & 6.87 & 7.63  & 1.29 & 0.27 & 0.30 & 0.71 & TarDAL~\cite{DBLP:conf/cvpr/LiuFHWLZL22}    & 6.32 & 7.42  & 1.04 & 0.27 & 0.22 & 0.58 \\
            SegMIF~\cite{Liu_2023_ICCV}    & 6.85 & {14.14} &\secondcolor{1.72} & 0.37 &\secondcolor{0.60} & 0.59 & SegMIF~\cite{Liu_2023_ICCV}    & 6.68 &\secondcolor{15.46} & 1.38 & 0.40 &\secondcolor{0.66} & 0.57 \\
            MURF~\cite{xu2023murf}      & 6.50 & 12.55 & 1.46 & 0.21 & 0.32 & 0.64 & MURF~\cite{xu2023murf}      & 6.13 & 15.08 & 0.96 & 0.21 & 0.31 & 0.57 \\
            EMMA~\cite{zhao2024equivariant}      & 6.92 & \firstcolor{15.23} & 1.49 &\secondcolor{0.38} & 0.59 & 0.69 & EMMA~\cite{zhao2024equivariant}      &\secondcolor{7.35} & 15.37 &\secondcolor{1.57} &\secondcolor{0.41} & 0.64 &\secondcolor{0.66} \\
            DCINN~\cite{wang2024general}     & 6.59 & 11.21 & 1.46 & 0.34 & 0.51 &\secondcolor{0.72} & DCINN~\cite{wang2024general}     & 6.98 & 13.34 & 1.43 & 0.38 & 0.52 & 0.64 \\
            MRFS~\cite{zhang2024mrfs}      &\secondcolor{6.94} & 12.07 & 1.26 & 0.34 & 0.55 & 0.70 & MRFS~\cite{zhang2024mrfs}      & 6.83 & 11.04 & 1.23 & 0.31 & 0.42 & 0.64 \\
            TIMFusion~\cite{liu2024task} & 6.75 & 12.31 & 1.37 & 0.35 & 0.53 & 0.70 & TIMFusion~\cite{liu2024task} & 6.58 & 13.52 & 1.14 & 0.33 & 0.46 & 0.64 \\
            TDFusion (Ours)  & \firstcolor{6.99} &\secondcolor{14.49} & \firstcolor{1.83} & \firstcolor{0.41} & \firstcolor{0.65} & \firstcolor{0.72} & TDFusion (Ours)  & \firstcolor{7.36} & \firstcolor{16.38} & \firstcolor{1.75} & \firstcolor{0.46} & \firstcolor{0.70} & \firstcolor{0.67}\\
            \bottomrule
	\end{tabular}}
\end{table*}%
\begin{figure*}[t]
	\centering

	\includegraphics[width=\linewidth]{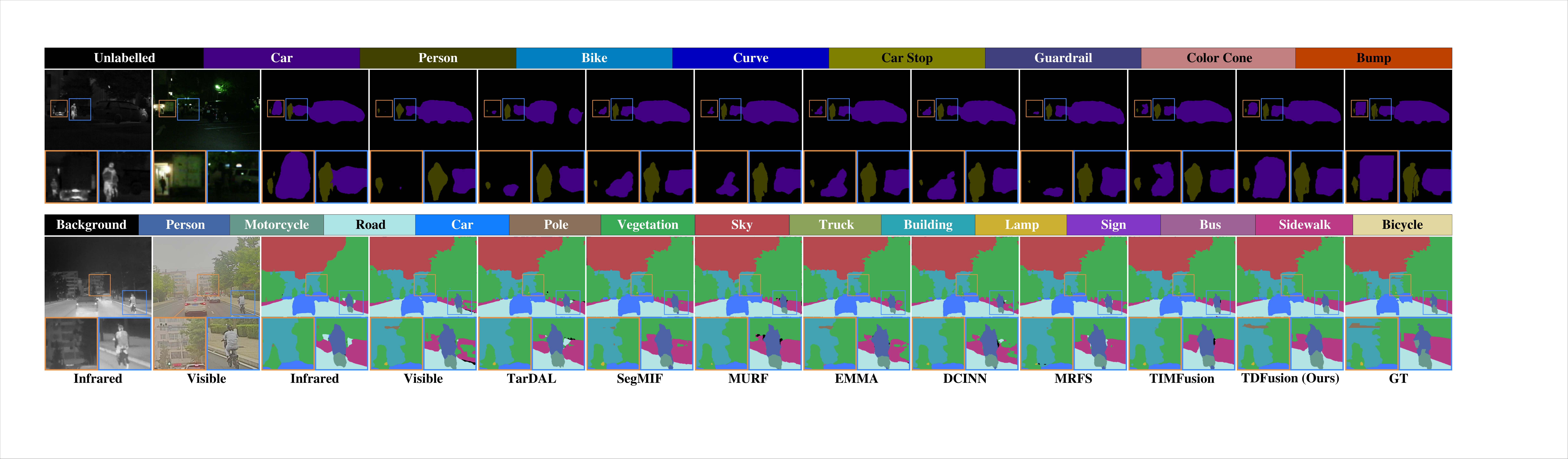}
        \vspace{-2em}
	\caption{{Visual comparison for Semantic Segmentation. The cases are “00726N” in MSRS dataset and “01438” in FMB dataset.}}
	\label{fig:SS}
	\vspace{-1.5em}
\end{figure*}
\cref{fig:fusion} presents a visual comparison of different methods. The fused images generated by TDFusion excel in preserving discriminative details, achieving balanced brightness, and maintaining clear object contours. 
It effectively preserves the target features from the infrared images and the background details from the visible images, resulting in fused images that are more natural and exhibit greater clarity across different environments.
These results highlight the advantages of TDFusion in detail preservation and visual performance.
More results are available in the supplementary material.
\cref{tab:fusion} presents the quantitative comparison of fusion over four datasets. TDFusion outperforms other methods across most metrics. This suggests that TDFusion not only enhances image details but also provides consistent fusion results across various scenarios. Compared to other methods, TDFusion shows superior adaptability and robustness, particularly in handling diverse image characteristics and challenging fusion tasks.

\subsection{Downstream Applications}

\begin{figure*}[t]
	\centering
	\includegraphics[width=\linewidth]{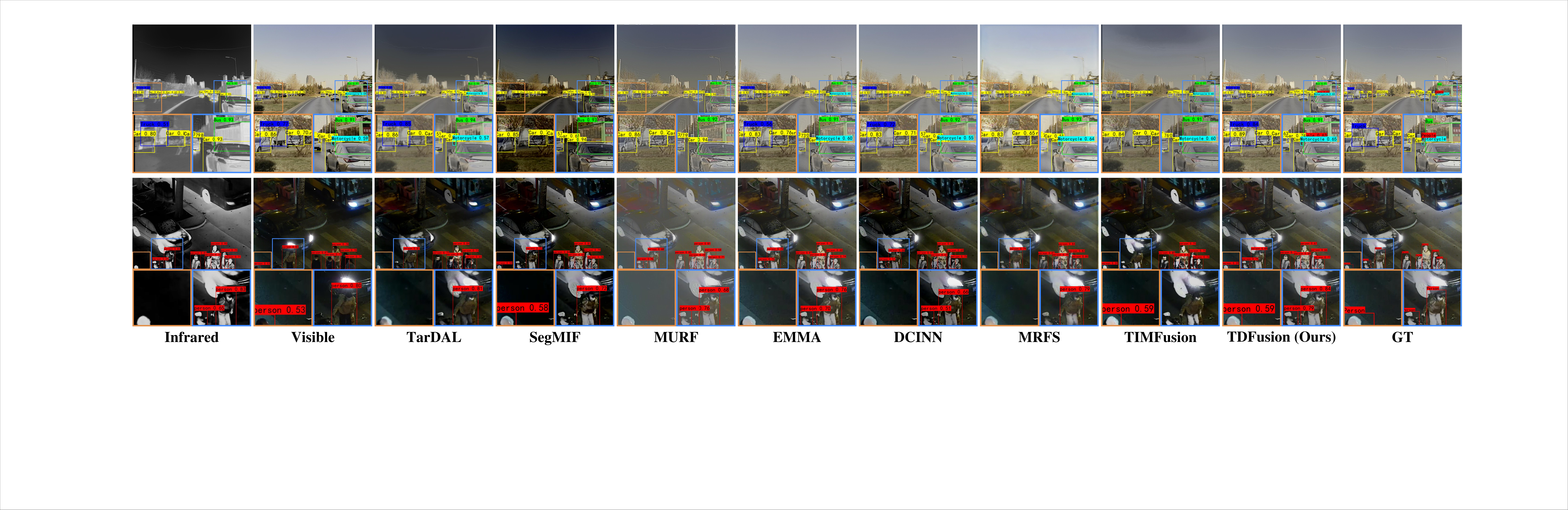}
        \vspace{-2em}
	\caption{{Visual comparison for Object Detection. The cases are “02236” in M3FD dataset and “210145” in LLVIP dataset.}}
	\label{fig:OD}
	\vspace{-1.5em}
\end{figure*}

This section validates the adaptability of fusion methods to downstream tasks. For a fair comparison, we adopt SegFormer~\cite{cheng2021segformer} and YOLOv8~\cite{ultralytics_yolov8_2023} as backbones and retrain the task networks for each fusion method over 300 epochs to evaluate their adaptability to semantic segmentation and object detection.
\cref{fig:SS} and \cref{fig:OD} present the visual comparisons of semantic segmentation and object detection, respectively. TDFusion outperforms in image detail retention, edge clarity, and object recognition, effectively identifying and segmenting objects. 
For semantic segmentation, the generated maps clearly distinguish different class regions, closely matching the ground truth. In object detection, the fused images exhibit more precise boundary localization for salient objects. This indicates that TDFusion maintains a better balance between fine details and overall context.
More results can be found in the supplementary material.

\begin{table}[h!]
\centering
\caption{Performance comparison of downstream applications. The \colorbox{firstcolor}{red} and \colorbox{secondcolor}{blue} markers represent the best and second-best.}
\label{tab:DS}
\vspace{-0.8em}
\resizebox{\linewidth}{!}{\begin{tabular}{lcccccccc}
\toprule
 & \multicolumn{4}{c}{\textbf{Semantic Segmentation}}                & \multicolumn{4}{c}{\textbf{Object Detection}}               \\ \cmidrule(lr){2-5}
\cmidrule(lr){6-9}
                                   & \multicolumn{2}{c}{\textbf{MSRS}} & \multicolumn{2}{c}{\textbf{FMB}} & \multicolumn{2}{c}{\textbf{M3FD}} & \multicolumn{2}{c}{\textbf{LLVIP}} \\ \cmidrule(lr){2-3} \cmidrule(lr){4-5} \cmidrule(lr){6-7} \cmidrule(lr){8-9}
                                  Methods & mAcc          & mIoU          & mAcc          & mIoU          & \small{mAP50}          & \small{mAP75}          & \small{AP50 }         & \small{AP75 }         \\ \midrule
Infrared                          & 83.23         & 69.49         & 58.85         & 51.98         & 79.12            & 53.05            & \firstcolor{96.03}           & \firstcolor{72.07}            \\ 
Visible                           & 83.44         & 73.76         & 65.12         & 57.96         & 82.21            & 54.82            & 91.78           & 48.66            \\ 
TarDAL                            & 81.93         & 71.35         & 62.86         & 55.33         & 83.16            & 56.39            & 93.79           & 62.71            \\ 
SegMIF                            & 85.73         & 74.25         &\secondcolor{65.97}         &\secondcolor{58.41}         & 83.61            &\secondcolor{58.23}            & 93.95           & 66.45            \\
MURF                              & 85.03         & 74.08         & 64.10         & 56.96         & 80.58            & 54.22            & 94.24           & 68.04            \\ 
EMMA                              &\secondcolor{85.99}         & 74.48         & 62.45         & 56.28         &\secondcolor{83.71}            & 56.91            & 94.00           & 66.21           \\ 
DCINN                             & 84.11         & 74.35         & 61.09         & 54.81         & 82.69            & 57.37            & 94.92           & 68.34            \\ 
MRFS                              & 84.76         &\secondcolor{74.50}         & 61.93         & 55.71         & 83.28            & 57.74            & 93.03           & 67.21            \\ 
TIMFusion                         & 83.67         & 73.58         & 63.70         & 57.24         & 83.22           & 56.08            & 93.76           & 61.33            \\ 
TDFusion                      & \firstcolor{86.04}         & \firstcolor{75.09}         & \firstcolor{67.17}         & \firstcolor{60.50}         & \firstcolor{86.27}            &  \firstcolor{59.71}            &\secondcolor{95.00}           &\secondcolor{69.18}            \\ \bottomrule
\end{tabular}}
\vspace{-2em}
\end{table}
\cref{tab:DS} shows the performance comparison across different methods in semantic segmentation and object detection. TDFusion outperforms other methods on most metrics, particularly in mIoU and mAP. This demonstrates that TDFusion effectively enhances the quality of fused images. It also improves the downstream task performance, especially in terms of accuracy and robustness in complex scenarios. Class-wise results are provided in the supplementary material.
\subsection{Task-driven Learnable Loss}

\begin{figure}[t]
	\centering
	\includegraphics[width=\linewidth]{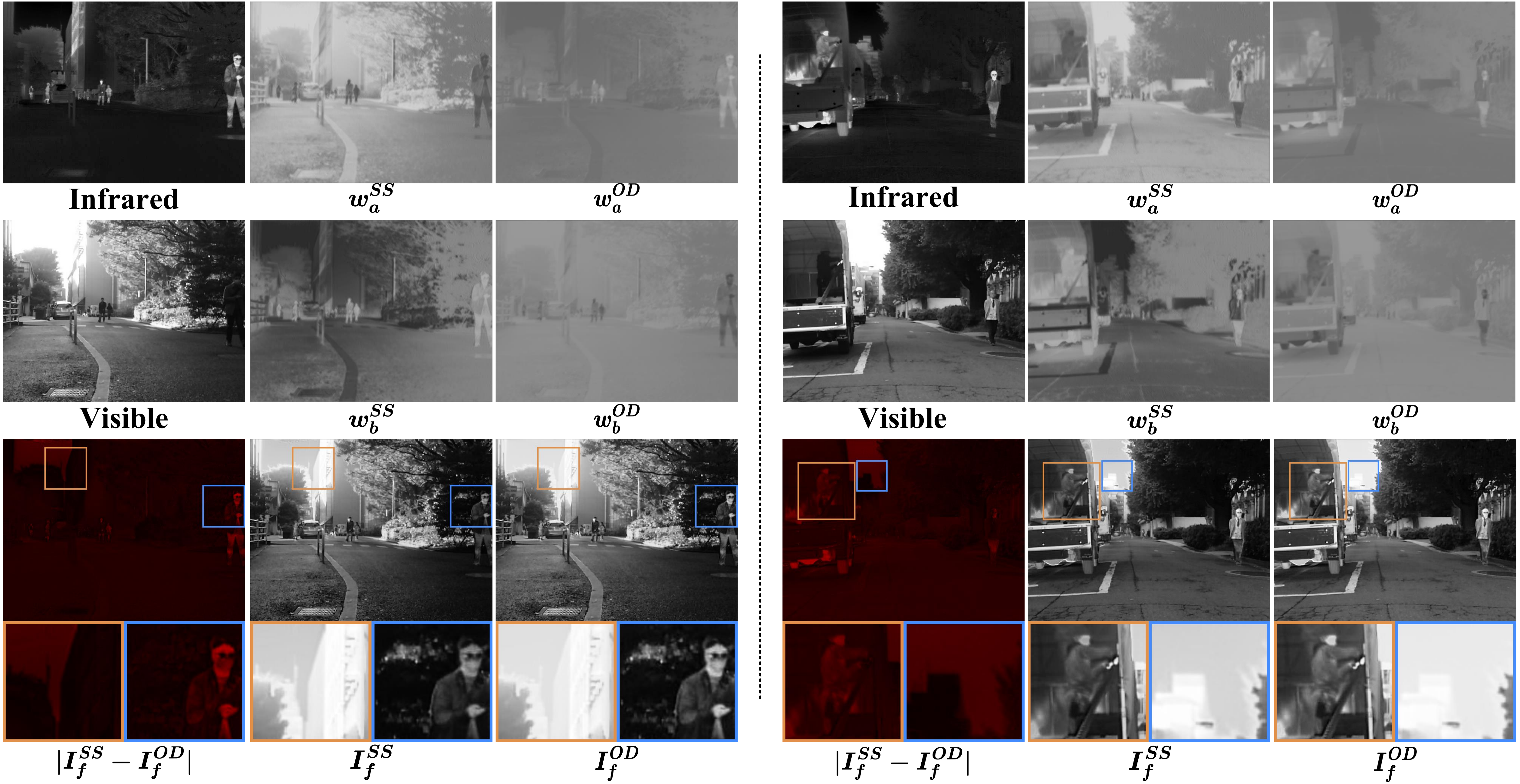}
        \vspace{-2em}
	\caption{{Visualisation of learnable loss for downstream tasks.}}
	\label{fig:LL}
	\vspace{-1.5em}
\end{figure}

Our framework incorporates a learnable fusion loss that models the preferences of downstream tasks for information from source images.
The models trained on FMB dataset and LLVIP dataset labeled as $SS$ and $OD$, are evaluated on MSRS dataset to simulate performance in unknown scenes, as shown in \cref{fig:LL}.
The results demonstrate that the fusion model adaptively selects information from infrared and visible images to satisfy task requirements.
In semantic segmentation, the model combines scene structure and texture, prioritizing boundaries. This improves segmentation under varying lighting conditions. Fusion weights $\{w^{SS}_a,w^{SS}_b\}$ indicate a preference for visible details and infrared advantages in low-light conditions.
In object detection, the model focuses on edge and contrast information, especially for instances like pedestrians and vehicles. Higher fusion weights are assigned to bright regions in infrared images, enhancing target detection in low-light conditions. Fusion weights $\{w^{OD}_a,w^{OD}_b\}$ reflect this preference.
Comparison of fusion losses across tasks reveals distinct differences, especially in highlighted regions, confirming that the model adapts to task-specific requirements by selecting the most relevant information from multimodal images.
{More results are provided in the supplementary material.}

\subsection{Ablation Studies}
To thoroughly evaluate the performance of our proposed algorithm, we conduct a series of ablation experiments on FMB dataset, and the detailed results are shown in \cref{tab:Ablation}. 
\begin{table}[h]
	\centering
	\centering

	\caption{Ablation experiment of fusion. The \colorbox{firstcolor}{red} denotes the best.}
	\label{tab:Ablation}
	\vspace{-0.8em}
	\resizebox{\linewidth}{!}{
		\begin{tabular}{cccccccc}
			\toprule
			\multicolumn{8}{c}{\textbf{Ablation Studies of fusion on FMB Dataset}}  
			\\
			&                       {Configurations}             & EN     & SF      & SCD    & VIF   & $Q^{AB/F}$   & SSIM   \\ \midrule
			\uppercase\expandafter{\romannumeral1} & fix $w_a$ and $w_b$ as $1/2$ & 6.60    & 13.73   & 1.58   & 0.39   & 0.60    & 0.72   \\
			\uppercase\expandafter{\romannumeral2} & w/o $\mathcal{L}_f^{grad}$ & 6.77   & 11.65   & 1.63   & 0.37   & 0.64   & 0.73   \\
			
			\uppercase\expandafter{\romannumeral3} &   $\theta_{\mathcal{F}}$ influenced by $\mathcal{L}_t$       & 6.80    & 13.85   & 1.70    & 0.41   & 0.66   & 0.73   \\
			\uppercase\expandafter{\romannumeral4} &              w/o Fusion learning              & {6.82}   & {14.07}   & {1.72}  & {0.41}   & {0.67}   & 0.72   \\
			
			\uppercase\expandafter{\romannumeral5} &              $I_f\!=\!w_a\!*\!I_a+w_b\!*\!I_b$                  & 6.75   & 11.49   & 1.65   & 0.38   & 0.62   & {0.73}   \\

			 \midrule

			&                            {Ours}                           &   \firstcolor{6.86} & \firstcolor{14.16} & \firstcolor{1.76} & \firstcolor{0.43} & \firstcolor{0.68} & \firstcolor{0.75} \\ \bottomrule
	\end{tabular}}
	\vspace{-1.5em}
\end{table}
In Exp.~\uppercase\expandafter{\romannumeral1}, we exclude the learnable fusion loss by fixing $w_a$ and $w_b$ to $1/2$. In Exp.~\uppercase\expandafter{\romannumeral2}, we omit the gradient loss from the loss function. In Exp.~\uppercase\expandafter{\romannumeral3}, we also allow the fusion module parameters to be jointly optimized by both the task loss $\mathcal{L}_t$ and fusion loss $\mathcal{L}_f$. In Exp.~\uppercase\expandafter{\romannumeral4}, we exclude the fusion module’s dedicated learning phase and update it during the outer update of the loss module, which tests the impact of the fusion module's training schedule. 
In Exp.~\uppercase\expandafter{\romannumeral5}, we replace our fusion method with a discriminative approach. The performance decline observed across different configurations confirms the rationality and effectiveness of our proposed method.
Visualization and more analysis are provided in the supplementary material.

\section{Conclusion}
To overcome the limitations of predefined fusion losses, which often fail to effectively guide the fusion process for downstream tasks, we propose a meta-learning-based framework for task-guided fusion.  
This framework includes a loss generation module that outputs the parameters of the learnable fusion loss.
The module is updated using a meta-learning approach, which alternates between inner and outer loop steps to enhance its ability to guide the fusion network.
Under varying fusion conditions, this module generates the optimal fusion loss for the downstream task. This enables the fusion network to produce fused images that minimize the task-specific loss.
The theoretical analysis explains how the downstream task loss guides the fusion loss in our framework.
Experiments using four publicly available fusion datasets and downstream tasks including semantic segmentation and object detection, demonstrate the effectiveness of our approach.
\section*{Acknowledgement}
This work has been supported by the National Natural Science Foundation of China under Grant 12201497 and 12371512.

{
    \small
    \bibliographystyle{ieeenat_fullname}
    \bibliography{refer}
}

\end{document}